\documentclass[runningheads]{llncs}
\usepackage{tabu}                     
\usepackage{multirow}                 
\usepackage{multicol}                 
\usepackage{multirow}                
\usepackage{float}                    
\usepackage{makecell}                 
\usepackage{booktabs}                 
\usepackage{graphicx}
\usepackage{subcaption}
\usepackage{algorithm}  
\usepackage{algpseudocode}
\usepackage{wrapfig}
\usepackage{float}
\usepackage{marvosym}

\begin{document}
\title{Self Adaptive Threshold Pseudo-labeling and Unreliable Sample Contrastive Loss for Semi-supervised Image Classification}
\titlerunning{STUC-SSIC}
%
%
\author{Xuerong Zhang\inst{1}
 \and Li Huang\inst{2}
\and Jing Lv\inst{1}\textsuperscript{(\Letter)} \and
Ming Yang} 
\authorrunning{X. Zhang et al.}
%
\institute{Nanjing Normal University, Nanjing, China \\
\email{05275@njnu.edu.cn
}\and
Jiangsu Open University, Nanjing, China}
\maketitle              

\begin{abstract}
Semi-supervised learning is attracting blooming attention, due to its success in combining unlabeled data. However, pseudo-labeling-based semi-supervised approaches suffer from two problems in image classification: (1) Existing methods might fail to adopt suitable thresholds since they either use a pre-defined/fixed threshold or an ad-hoc threshold adjusting scheme, resulting in inferior performance and slow convergence. (2) Discarding unlabeled data with confidence below the thresholds results in the loss of discriminating information. To solve these issues, we develop an effective method to make sufficient use of unlabeled data. Specifically, we design a self adaptive threshold pseudo-labeling strategy, which thresholds for each class can be dynamically adjusted to increase the number of reliable samples. Meanwhile, in order to effectively utilise unlabeled data with confidence below the thresholds, we propose an unreliable sample contrastive loss to mine the discriminative information in low-confidence samples by learning the similarities and differences between sample features. We evaluate our method on several classification benchmarks under partially labeled settings and demonstrate its superiority over the other approaches.
\end{abstract}
\section{Introduction}
In the past several years, our computer vision community has witnessed inspiring progress, thanks to fast developments of deep learning~\cite{1}. Such a big success is closely dependent on supervised training with sufficient labeled data. However, a large amount of labeled data are usually laborious and expensive to obtain. To mitigate the demand for labeled data, Semi-supervised learning(SSL)~\cite{4,7} has been proposed as a powerful approach to leverage unlabeled data.
\par Consistency regularization~\cite{8,9,10} and pseudo labeling~\cite{11,12} are two powerful techniques in modern SSL, where SSL methods based on pseudo-labeling have achieved good performance. Briefly, the method first train the model based on the labeled data and then use the model’s predictions on unlabeled data as pseudo-labels. It is obvious that predictions on unlabeled data are not reliable. If the model is iteratively trained with incorrect pseudo labels, it will suffer the confirmation bias issue~\cite{13}. To address this dilemma, Sohn et al.~\cite{4}  proposed FixMatch, which simply set a fixed confidence threshold to discard potentially unreliable samples.While this strategy can make sure that only high-quality unlabeled data contribute to the model training, it also incurs a low utilization of the whole unlabeled set. Xu et al. proposed Dash~\cite{14},  which proposes to gradually grow the fixed global threshold as the training progresses. Although the utilization of unlabeled data is improved, their ad-hoc threshold adjusting scheme is arbitrarily controlled by hyper-parameters. Zhang et al.~\cite{15} proposed the curriculum pseudo-labeling method to flexibly adjust the thresholds and achieve best result in several SSL benchmark testing datasets.
\par However, these pseudo-labeling SSL methods  alleviate confirmation bias to some extent, but still have many problems and can be summarized in two key aspects: (1) The dynamic thresholding strategies, while taking into account the learning difficulties of different classes, are still mapped from predefined fixed global thresholds and are not sufficient to adjust the thresholds according to the actual learning progress of the models. (2) These methods discards low-confidence pseudo labels and does not fully utilize the unlabeled data and consumes a longer training time.
\par To solve the above issues, a novel method for semi-supervised image classification is proposed, we call our method as \textbf{S}elf Adaptive \textbf{T}hreshold Pseudo-labeling and \textbf{U}nreliable Sample \textbf{C}ontrastive Loss for \textbf{S}emi-\textbf{s}upervised \textbf{I}mage \textbf{C}lassification ( STUC-SSIC), which consists of self-adaptive threshold pseudo-labeling (SATPL) and unreliable sample contrastive loss (USCL). SATPL proposes that different datasets have different global thresholds, and each class has its own local threshold based on the dynamic global threshold. The number of reliable samples is gradually increased by adjusting the threshold for each class in an adaptive manner. At the same time, we believe that unlabeled data below the thresholds can provide discriminating information to the model. Unlike previous methods~\cite{4,14}, which discard unlabeled data below the thresholds, we construct an unreliable sample contrastive loss (USCL) to mine discriminating information in the unreliable samples and improve the utilization of the unlabeled dataset. Experimental results show that STUC-SSIC improves the utilization of unlabeled data and improves the classification performance on semi-supervised image classification tasks on three datasets, including CIFAR-10, CIFAR-100 and STL-10.
\par In summary, the following are the main contributions of our paper:
\begin{itemize}
    \item[$\bullet$]  We design a self-adaptive threshold pseudo-labeling (SATPL) strategy, considering that the model performance gradually improves with the training process, the threshold of each class is adjusted in an adaptive manner to gradually increase the number of reliable samples. 
    \item[$\bullet$] An unreliable sample contrastive loss (USCL) is proposed, which can effectively mines the discriminative information in low-confidence samples by learning the similarities and differences between sample features, enabling the model to utilize all unlabeled samples and speeding up the convergence of the model.
    \item[$\bullet$] Experimental results on the CIFAR-10, CIFAR-100 and STL-10 datasets show the superiority of our method.
\end{itemize}
\section{Related work}
\subsection{Confidence-based Pseudo-labeling SSL}
Semi-supervised learning has been researched for decades, and the essential idea is to learn from the unlabeled data to enhance the training process. In semi-supervised learning, a confidence-based strategy has been widely used along with pseudo labeling so that the unlabeled data are used only when the predictions are sufficiently confident. Sohn et al.~\cite{4}  proposed FixMatch, which simply set a fixed confidence threshold to discard potentially unreliable samples. While this strategy can make sure that only high-quality unlabeled data contribute to the model training, it also incurs a low utilization of the whole unlabeled set. Xu et al.~\cite{14}  proposed Dash,  which proposes to gradually grow the fixed global threshold as the training progresses. Although the utilization of unlabeled data is improved, their ad-hoc threshold adjusting scheme is arbitrarily controlled by hyper-parameters. Zhang et al.~\cite{15} proposed the curriculum pseudo-labeling method to flexibly adjust the thresholds and achieve best result in several SSL benchmark testing datasets. However, these methods do not use the full unlabeled data for model training.
\subsection{Contrastive learning}
Recent contrastive learning studies have presented promising results to directly leverage the unlabeled data. Such methods are to train a representation learning model by automatically constructing similar and dissimilar instances, which essentially encourage similar feature representations between two random crops from the same image and distinct representations among different images. SimCLR~\cite{21,22} shows that image augmentation, nonlinear projection head and large batch size plays a critical role in contrastive learning. Since large batch size usually requires a lot of GPU memory, which is not very friendly to most of researchers. MoCo~\cite{23} proposed a momentum contrast mechanism that forces the query encoder to learn the representation from a slowly progressing key encoder and maintain a memory buffer to store a large number of negative samples. In Corporate relative valuation(CRV), Yang et al.~\cite{39} proposed
HM2, which adopted additional triplet loss with embedding of competitors as
the constraint to learn more discriminative features. Consequently, HM2 can explore
 company intrinsic properties to improve CRV. In Class-Incremental Learning
 (CIL), Yang et al.~\cite{40} proposed a semi-supervised style Class Incremental
Learning without Forgetting (CILF) method, designs to regularize classiffcation
with decoupled prototype based loss, which can improve the intra-class and inter-class
 structure signiffcantly, and acquires a compact embedding representation for novel class detection in result. In Semi-Supervised Image Captioning, Yang
et al.~\cite{41} developed a novel relation consistency between augmented images and
corresponding generated sentences to retain the important relational knowledge.
In Course Recommendation System, Yang et al. ~\cite{42} applied contrastive learning
 to learn effective representations of talents and courses. Especially in an open environment, Yang et al.~\cite{43} proposed to introduce a sampling mechanism to actively select valuable out-of distribution(OOD) instances, balancing pseudo-ID and pseudo-OOD instances to enhance ID classifiers and OOD detectors. Due to the superior performance of contrastive learning at directly exploiting unlabelled samples, we joint contrastive learning to use unlabeled data below the thresholds to increase the model’s performance. Also in this paper, we use the idea of how to select negative samples in contrastive learning.
\subsection{Self-supervision in Semi-supervised Learning}
Many recent state-of-the-art semi-supervised learning methods adopt the self-supervised representation learning methods~\cite{22,25} to jointly learn good feature representation, which are similar considerations to our method. Specifically, S4L~\cite{25} integrated two pretext-based self-supervised approaches in SSL and showed that unsupervised representation learning complements existing SSL methods. CoMatch~\cite{26} proposed graph-based contrastive learning to learn better representations. LaSSL~\cite{27} enjoys benefits from exploring wider sample relations and more label information, through injecting class-aware contrastive learning and label propagation into the standard self-training. However, methods with instance discrimination~\cite{21,23}, which treats each image as its own class, can hurt semi-supervised image classification as it partially conflicts with it.
\section{Methods}
In this section, we will first revisit the preliminary work on semi-supervised learning. Then, we will introduce our proposed STUC-SSIC framework. After that, the algorithm and the implementation details will also be explained.
\subsection{Preliminaries on Semi-supervised Learning}
For a semi-supervised image classification problem, where  we train a model using $M$ labeled samples and $N$ unlabeled samples, where $N \gg M$. We use mini-batches of labeled instances, $\mathcal{X} = \{(x_{i},y_{i})\}_{i=1}^{B}$ and unlabeled instances, $\mathcal{U} = \{(u_{i})\}_{i=1}^{\mu B}$, where the scalar $\mu$ denotes the ratio of unlabeled samples to labeled samples in a batch, and $y$ is is the one-hot vector of the class label $c\in\{1,2,...,C\}$.  The goal of semi-supervised learning is to use the dataset to train a model with parameter $\theta$, which consists of an  encoder network $f$ and a softmax classifier $g$ to produce a distribution over classes $p=f(g(\cdot))$~\cite{27}. Referring to FixMatch~\cite{4}, We apply weak augmentations $\omega(\cdot)$ on all images and an additional strong augmentation $\Omega(\cdot)$ only on unlabeled samples, which can enhance the model's ability to adapt to changes in the data. The supervised loss for labeled data is:\\
\begin{equation}
    \mathcal{L}_{s}=\frac{1}{B}\sum_{i=1}^{B}H(y_{i},p_{\theta}(y|\omega(x_{i})))
\end{equation}
where $B$ is the batch size, $H(\cdot,\cdot)$ refers to cross-entropy loss,  $p_{\theta}(\cdot)$ is the output probability from the model.
\subsection{STUC-SSIC}
\begin{figure}[t]
    \centering
    \includegraphics[width=0.95\linewidth]{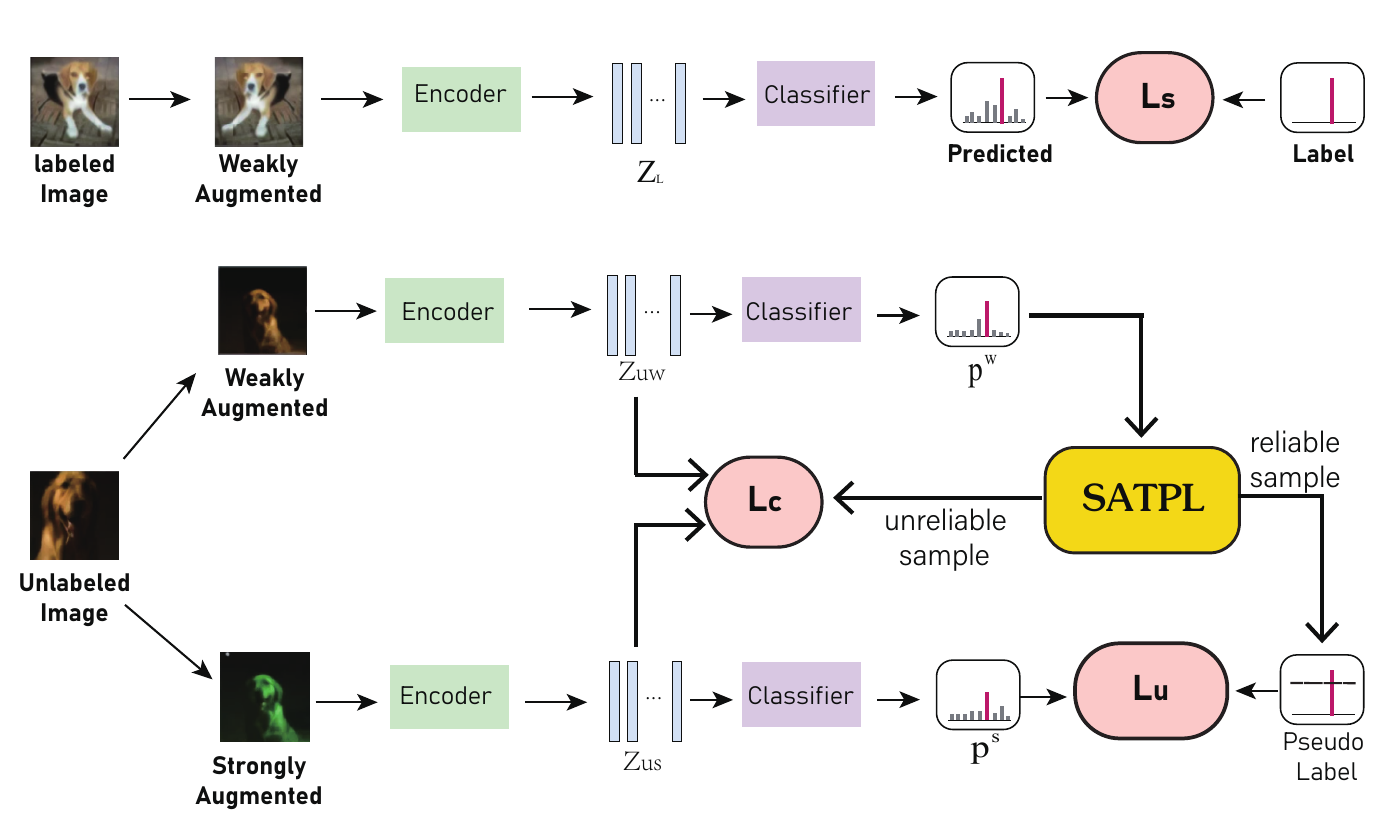}
    \caption{Illustration of STUC-SSIC.
Weak data augmentation labeled data (top) with true labels constitutes a supervised loss.For unlabeled data (bottom), self-adaptive threshold pseudo-labeling (SATPL) generate the current local threshold, pseudo-labels are generated only when the class probability of weak data augmented samples is higher than the local threshold. The prediction of strong data augmented samples with pseudo-labels constitute an unsupervised loss. Unreliable samples (samples below the thresholds)  construct a new contrastive loss for training.}
    \label{模型图}
\end{figure}

In order to solve the problems of inability to fully utilize unlabeled data and low quality of pseudo-labels in deep semi-supervised image classification, we propose  STUC-SSIC method, and the overall architecture is shown in Fig.~\ref{模型图}. It mainly consists of two parts: self-adaptive threshold pseudo-labeling (SATPL) and unreliable sample contrastive loss (USCL). SATPL can generate reliable pseudo-labels, and USCL can improve the utilization of unlabeled data, and the combination of the two can improve both the classification accuracy and the convergence speed of the model.
\subsubsection{Self Adaptive Threshold Pseudo-labeling.}
The pseudo-labeling-based SSL method has the following drawbacks: due to the high fixed threshold, most of the samples cannot be added to the model training, which prevents the model from identifying classes that are difficult to learn. In addition, some recent approaches~\cite{14,15} have proposed different strategies for screening samples with dynamically adjusted thresholds. Although those strategies take into account the degree of learning difficulty of different classes, are still mapped from predefined fixed global thresholds, which are not sufficient to adjust the thresholds according to the actual learning progress of the model. Therefore, we introduce SATPL, which global thresholds are varied according to different datasets, and local thresholds are scaled on top of global thresholds.
\par We propose that at the beginning of training, when the model's learning ability is weak, a lower global threshold is needed to utilize more unlabeled data thus accelerating the model's convergence. As training progresses and the model's learning ability becomes stronger, a higher global threshold is needed to filter out erroneous pseudo-labels and mitigate the confirmation bias in order to improve the quality of pseudo-labels. In addition, since each class has different learning difficulties and different confidence predictions for the samples, the local threshold for each class is dynamically adjusted during the training process to allow the learning-difficulty classes to produce more pseudo-labels, to keep the pseudo-labels' classes balanced, and to alleviate confirmation bias issues.
\par First define the global thresholds of the model for different datasets. Inspired by FreeMatch \cite{16}, we estimate the global confidence as the exponential moving average (EMA) of the confidence at each training time step. We initialize $\tau_{t}$ as $\frac{1}{C}$ where C indicates the number of classes. The global threshold  $\tau_{t}$ is defined and adjusted as:\\
\begin{equation}
\tau_{t} = \left\{
        \begin{array}{lr}
             \frac{1}{C}, & t=0 \\
             \lambda \tau_{t-1} + (1-\lambda)\frac{1}{\mu B}\sum_{i=1}^{\mu B}{\rm max}(p_{\theta,t}(y|\omega(u_{i}))),  &  otherwise
        \end{array}
        \right.
\end{equation}\\
where $t$ denotes the $t$-th iteration of model training, $\lambda \in (0,1)$ is the momentum decay of EMA.
\par Then local thresholds for each class are obtained by scaling the global threshold using the learning effect of the current class. The learning status of the model for each class during training can be calculated as follows:\\
\begin{equation}
    \phi_{t}(c) = \sum_{i=1}^{N}{\rm I}({\rm max}(p_{\theta,t}(y|\omega(u_{i})))>\tau_{t}){\rm I}({\rm argmax}(p_{\theta,t}(y|\omega(u_{i})))=c)
\end{equation}
where $ \phi_{t}(c)$ denotes the learning effect of the current class, which can be reflected by the number of samples that are predicted to fall into this class and above the global threshold at round $t$ of training.
Larger $ \phi_{t}(c)$ indicates a better estimated learning effect. By applying the following normalization to  $ \phi_{t}(c)$ to make its range between 0 to 1, it can then be used to scale the local threshold $\sigma_{t}(c)$ :
\begin{equation}
    \Phi_{t}(c) = \frac{\phi_{t}(c)}{\max\limits_{c}(\phi_{t})}
\end{equation}
\begin{equation}
    \sigma_{t}(c)=\Phi_{t}(c)\cdot\tau_{t}
\end{equation}
\par Thus, the unsupervised loss in STUC-SSIC can be expressed as:
\begin{equation}
    \mathcal{L}_{u} = \frac{1}{\mu B}\sum_{i=1}^{\mu B}{\rm I}({\rm max}(q_{b})\geq\sigma_{t}({\rm argmax}(q_{b}))) H(\hat{q}_{b},p_{\theta}(y|\Omega(u_{i})))
\end{equation}
where $q_{b}=p_{\theta}(y|\omega(u_{i}))$, $\hat{q}_b$ is the pseudo-label, and H is the cross-entropy loss function.
\subsubsection{Unreliable Sample Contrastive loss.}
The key to contrastive learning is to construct positive and negative instance sample pairs.  Most  existing contrastive learning methods~\cite{21,23} focus on instance-level information, which to separate different (negative) data pairs while aggregating similar (positive) data. However, such method will inevitably cause class collision problems, which hurts the quality of the learned representation. Moreover, this contrast loss of treating an instance image as a class introduced into a semi-supervised image classification algorithm would contradict the semi-supervised classification task and affect the performance of the classification.
\par Therefore, we construct a contrastive loss function suitable for semi-supervised image classification tasks, and the expression is as follows:
\begin{equation}
    \mathcal{L}_{c}=\frac{1}{\mu B}\sum_{i=1}^{\mu B}-\xi(u_{i}) {\rm log}\frac{e^{{\rm sim}(Z_{i},Z_{i}^{+})/T}}{e^{{\rm sim}(Z_{i},Z_{i}^{+})}+\sum_{u_{i}\in N(u_{b})}e^{{\rm sim}(Z_{j},Z_{j}^{-})/T}}
\end{equation}
where $\xi(u_{i})={ I}({\rm max}(q_{b})<\sigma_{t}({\rm argmax}(q_{b})))$ denotes select unreliable samples below the threshold for training. $Z_{i}$ is the feature representation of sample, $Z_{i}^{+}$  is the feature representation of positive sample, $N(u_{b})$ is the set of negative samples, obtained by randomly sampling uniformly from dissimilar negative samples, $Z_{j}^{-}$ is the feature representation of negative sample, ${\rm sim}(\cdot,\cdot)$ denotes the cosine similarity function, $T$ is the temperature hyper-parameters, these parameters are set from the design of the InfoNCE~\cite{23} loss.
\par Given an unlabeled  image $u_{i}$ in a batch of samples,weak and strong  augmented views can be obtained by $\omega(u_{i}),\Omega(u_{i})$ respectively, and calculate the corresponds embedding $Z_{uw}^{i}=f(\omega(u_{i})),Z_{us}^{i}=f(\Omega(u_{i}))$. Then, we calculate the similarities between the instances of the weakly augmented images and other instance image. A softmax layer can be adopted to process the calculated similarities, which then produces a relationship distribution:
\begin{equation}
    \gamma_{w}^{i}=\frac{e^{{\rm sim}(Z_{uw}^{i},Z^{j})/T_{w}}}{\sum_{k=1}^{K}e^{{\rm sim}(Z_{uw}^{i},Z^{k})/T_{w}}}
\end{equation}
\begin{equation}\gamma_{s}^{i}=\frac{e^{{\rm sim}(Z_{us}^{i},Z^{j})/T_{s}}}{\sum_{k=1}^{K}e^{{\rm sim}(Z_{us}^{i},Z^{k})/T_{s}}}
\end{equation}
where $T_{w}$ and $T_{s}$ is the temperature parameter, which is used to control the clarity of the distribution. The smaller the temperature coefficient, the clearer the distribution, highlighting the most similar features in the distribution. In order to save parametric quantities, $T_{w}$ and $T_{s}$ are taken to be 0.1 in the experiments.\\
When a sample is similar to a sample with both strong  augmentations as well as weak augmentations, this sample is considered as a positive sample, which added to the set of positive samples, and the rest of the samples are regarded as negative samples. The positive sample was selected as follows:
\begin{equation}
    \Lambda=\{Z_{i}|\gamma_{w}^{i}>\varepsilon_{1},\gamma_{s}^{i}>\varepsilon_{2}\}
\end{equation}
which $\varepsilon_{1}$ and $\varepsilon_{2}$ are thresholds for filtering similarity. In general, the strong  augmentation image is too different from the original image, so the threshold is set smaller than  the threshold of weak  augmentation image. Then, calculate the mean of all positive samples in the set as the final positive sample:
\begin{equation}
    Z_{i}^{+}=\frac{1}{|\Lambda|}\sum_{Z^{i}\in\Lambda}Z^{i}
\end{equation}
\begin{algorithm}[t]
  \caption{STUC-SSIC algorithm} 
  \begin{algorithmic}[1]
    \Require
      Batch of labeled samples $\mathcal{X}=\{(x_b,y_b)\}_{b=1}^{B}$,batch of unlabeled samples  $\mathcal{U}=\{(u_b)\}_{b=1}^{\mu B}$,  weak augmentation strategy $\omega(\cdot)$, strong augmentation strategy $\Omega(\cdot)$, total number of iterations $T_t$, the number of class $C$;
    \Ensure Parameters of the image classification model $\theta$.
			\State Initialize global thresholds $\tau_0=\frac{1}{C}$, local thresholds $\sigma_{t}(c) = \emptyset$;
			\For{$t$  to  $T_t$}
			\For{$x_b \in \chi$}
			\State $Z_L=f(\omega(x_b))$; //Extracting features of the labeled samples
			\State $p_{\theta}(y \mid \omega(x_b))=g(f(\omega(x_b)))$; //Getting the predicted probabilities
			\EndFor
			\State Calculate the supervised loss $\mathcal{L}_{s}$ via Eq.(1);
			\For{$c=1$ to $C$}
			\State Calculate the learning status of each class $\phi_{t}(c)$ via Eq.(3);
			\State Calculate the normalization of learning status $\Phi_{t}(c)$ via Eq.(4);
			\State Calculate local thresholds $\sigma_{t}(c)$ via Eq.(5);
			\EndFor
			\For{$u_b \in u$} 
			\State $Z_{uw}=f(\omega(u_b))$, $Z_{us}=f(\Omega(u_b))$; //Extracting features
			\State $p_{\theta}(y \mid \omega(u_b))=g(f(\omega(u_b)))$, $p_{\theta}(y \mid \Omega(u_b))=g(f(\Omega(u_b)))$;
			\If{{\rm max}$(p_{\theta}(y \mid \omega(u_b)))>\sigma_{t}(c)$}
			\State $\hat{q_{b}}={\rm argmax}(p_{\theta}(y \mid \omega(u_b)))$; //Getting Pseudo Labels
			\Else
			\State The set of positive samples and negative samples are obtained via Eq.(8), Eq.(9), Eq.(10);
			\EndIf
			\EndFor
			\State Update global thresholds $\tau_t$ via Eq.(2);
			\State Calculate  the unsupervised loss  $\mathcal{L}_{u}$ via Eq.(6);
			\State Positive samples in contrastive loss are obtained via Eq.(11);
			\State Calculate the contrastive loss $\mathcal{L}_{c}$ via Eq.(7);
			\State Calculate total loss $\mathcal{L}$ via Eq.(12);
			\EndFor
  \end{algorithmic}
\end{algorithm}
\subsection{Overall loss}
In summary, the total loss at each mini-batch is:
\begin{equation}
    \mathcal{L}=\mathcal{L}_{s}+\lambda_{u}\mathcal{L}_{u}+\lambda_{c}\mathcal{L}_{c}
\end{equation}
where $\mathcal{L}_{s}$ is the supervised loss on labeled data, $\lambda_{u}$ and $\lambda_{c}$ are two weight parameter for the unsupervised loss and the unreliable sample contrastive loss respectively. We follow the setting of FixMatch~\cite{4}, set  $\lambda_{u}$ to 1.0. We adjust $\lambda_{c}$ in an exponentially ramping down manner:\\
\begin{equation}
    \lambda_{c} = \left\{
        \begin{array}{lr}
             \lambda_{c}^{0}, & t=0 \\
             \lambda_{c}^{0}e^{-\frac{t}{T_{t}}},  &  otherwise
        \end{array}
        \right.
\end{equation}
which $\lambda_{c}^{0}$ is set as the maximum value of $\lambda_{c}$, and $T_{t}$ is the total number of training iterations. The purpose of USCL is to learn a better representation of the features and is not related to the downstream task. Early in training, the contrastive loss can make a relatively large contribution to the model. As the training progresses to the later stages of training, the model should focus more on the downstream classification task, so it will be decreasing and training to the later stages focuses more on the classification task. We present the procedure of  STUC-SSIC in Algorithm 1.

\section{Experiments}
We evaluated the STUC-SSIC and other semi-supervised algorithms on three public datasets. In addition, we studied STUC-SSIC’s performance with varying ratios of labeled data to highlight the benefits of including unlabeled data in training. Finally, we conducted an extensive ablation experiment to validate the efficacy of each component of our method.
\subsection{Datasets and experimental setup}
\subsubsection{Small Dataset.} CIFAR-10~\cite{17} and CIFAR-100~\cite{17}. The CIFAR-10 dataset consists of 60000 32 $\times$32 colour images in 10 classes, with 6000 images per class. There are 50000 training images and 10000 test images. CIFAR-100 is just like the CIFAR-10, except it has 100 classes containing 600 images each. There are 500 training images and 100 testing images per class.
\subsubsection{Medium Dataset.}  STL-10~\cite{18}. The images in STL-10 are from ImageNet, and there are 113,000 color images of 96$\times$96 resolution, with 5,000 labeled training samples and 8,000 labeled test samples, and 100k unlabeled color images are also provided. There are some samples in these unlabeled color images with different classes than the training set, making this dataset a challenging task.
\subsubsection{Implementation Details.} Following FixMatch~\cite{4}, for CIFAR-10 and CIFAR-100, we adopt WideResNet-28-2 and WideResNet-28-8 \cite{19} as the backbone, respectively, while using Wide ResNet-37-2~\cite{20} for STL-10. The batch size of the labeled sample dataset $B$ is set to 64, and the ratio of unlabeled samples  to labeled samples $\mu$ is set to 7. The unsupervised loss $\lambda_{u}$ is set to 1. The temperature coefficient in the contrastive loss term is set to 0.1. For the CIFAR-10 dataset 800k iterations of training were performed, for the CIFAR-100 dataset 300k iterations of training were performed, and for the STL-10 dataset 500k iterations of training were performed. We use an exponential moving average with a decay rate of 0.999 to test our model and repeat the same experiment for five runs with different seeds to report the mean accuracy.
\subsection{Comparison Experiment Results }
In Table 1, we compare the testing accuracy of our proposed method against recent SSL methods with a varying number of labeled samples. These  results demonstrate that STUC-SSIC achieves the best performance on CIFAR-10 and STL10 datasets, and it produces very close results on CIFAR-100 to the best competitor.
\begin{table}[t]
    \renewcommand{\arraystretch}{1.3}
    \centering
    \label{t1}
    \caption{Comparison of our STUC-SSIC to other relevant works. Top-1 testing accuracy (\%) for CIFAR-10, CIFAR-100 and SVHN on 5 different folds. Each result is reported as the average of 5 runs.Bold indicates the best result.}
    \resizebox{1\textwidth}{18mm}{
    \begin{tabular}{c|ccc|ccc|ccc}
        \Xhline{1pt}
        Dataset & \multicolumn{3}{c}{CIFAR-10}  & \multicolumn{3}{c}{CIFAR-100} & \multicolumn{3}{c}{STL-10}\\ 
        \Xcline{1-1}{0.4pt}
        \Xhline{1pt}
         Label Amount & 40 & 250 & 4000 & 400 & 2500 & 10000 & 40 & 250 & 1000 \\
        \Xcline{1-1}{0.4pt}
        \Xhline{1pt}
        Mean-Teacher\cite{38} & 40.38\scalebox{0.7}{±3.21} & 62.30\scalebox{0.7}{±0.78} & 83.18\scalebox{0.7}{±0.12}  & 12.35\scalebox{0.7}{±1.63} & 46.40\scalebox{0.7}{±0.48} & 54.18\scalebox{0.7}{±0.11} & 28.36\scalebox{0.7}{±2.26} & 50.34\scalebox{0.7}{±1.21} & 53.17\scalebox{0.7}{±1.08}\\
        MixMatch\cite{35} & 50.98\scalebox{0.7}{±4.23} & 83.80\scalebox{0.7}{±0.45} & 86.71\scalebox{0.7}{±0.09}  & 25.64\scalebox{0.7}{±1.23} & 50.80\scalebox{0.7}{±0.43} & 61.71\scalebox{0.7}{±0.19} & 37.28\scalebox{0.7}{±3.19} & 62.50\scalebox{0.7}{±0.56} & 64.31\scalebox{0.7}{±0.09} \\
        ReMixMatch\cite{34} & 76.84\scalebox{0.7}{±3.21} & 88.98\scalebox{0.7}{±0.08} & 90.76\scalebox{0.7}{±0.13}  & 44.82\scalebox{0.7}{±1.62} & 62.98\scalebox{0.7}{±0.18} & 66.96\scalebox{0.7}{±0.23} & 40.89\scalebox{0.7}{±2.04} & 66.38\scalebox{0.7}{±0.09} & 69.74\scalebox{0.7}{±0.12} \\
        FixMatch\cite{4} & 85.90\scalebox{0.7}{±3.01} & 89.09\scalebox{0.7}{±0.62} & 91.26\scalebox{0.7}{±0.04}  & 40.21\scalebox{0.7}{±2.01} & 61.71\scalebox{0.7}{±0.11} & 66.26\scalebox{0.7}{±0.16} & 42.45\scalebox{0.7}{±3.11} & 68.02\scalebox{0.7}{±0.13} & 71.26\scalebox{0.7}{±0.07} \\
        FlexMatch\cite{15}& 88.26\scalebox{0.7}{±0.06} & 88.95\scalebox{0.7}{±0.66} & 91.38\scalebox{0.7}{±1.01}  & 45.52\scalebox{0.7}{±1.22} & 63.11\scalebox{0.7}{±0.19} & 67.01\scalebox{0.7}{±0.12} & 47.23\scalebox{0.7}{±2.62} & 68.82\scalebox{0.7}{±0.19} & 72.63\scalebox{0.7}{±0.22} \\
        FullMatch\cite{31} & 89.77\scalebox{0.7}{±3.21} & 91.76\scalebox{0.7}{±0.42} & 92.82\scalebox{0.7}{±1.12}  & 45.19\scalebox{0.7}{±0.51} & 65.37\scalebox{0.7}{±0.27} & 68.46\scalebox{0.7}{±0.18} & 52.18\scalebox{0.7}{±0.66} & 72.11\scalebox{0.7}{±0.11} & 73.99\scalebox{0.7}{±0.21} \\
        \Xcline{1-1}{0.4pt}
        \Xhline{1pt}
        \textbf{STUC-SSIC} & \textbf{91.89}\scalebox{0.7}{±1.06} & \textbf{93.09}\scalebox{0.7}{±0.06} & \textbf{94.21}\scalebox{0.7}{±0.09}  & \textbf{46.88}\scalebox{0.7}{±0.21} & 65.39\scalebox{0.7}{±0.19} & 68.55\scalebox{0.7}{±0.15} & \textbf{55.89}\scalebox{0.7}{±0.32} & \textbf{73.08}\scalebox{0.7}{±0.09} & \textbf{75.00}\scalebox{0.7}{±0.18} \\
        \Xhline{1pt}
    \end{tabular}}
\end{table}

 On CIFAR-10 task with 40 labels, our method achieves the mean accuracy of 91.89\%, which outperforms the FixMatch~\cite{4} method by 6\%. This is mainly due to the fact that FixMatch uses a fixed threshold to filter the samples, and cannot dynamically change the threshold according to the learning state of the model. On CIFAR-100 task with 40 labels, our method achieves the mean accuracy of 46.88\%, which outperforms the FullMatch~\cite{31} and FixMatch~\cite{4} method by 1\% and 6\% , respectively. And with a high number of labels, the accuracy of the our proposed  method can be slightly higher than the FullMatch~\cite{31} method. On a complex dataset like STL-10, our method achieves good classification results, which  may be attributed to the fact that the self-supervised loss component gives the model additional supplementary information, allowing the model to obtain additional self-supervised information when it lacks supervisory information in the case of very few labels thus allowing the model's performance to be improved. In addition the design of the local dynamic thresholds for each class allows the model to produce more balanced pseudo-labels, resulting in improved classification performance.
\par \begin{wraptable}{r}{0.4\textwidth}
    \centering
    \label{tbl:table2}
    \vspace{-9mm}
    \caption{Top-1 testing accuracy (\%) and runtime (sec./iter.) on CIFAR-100 with 400 labels.}
    \begin{tabular}{c|cc}
        \Xhline{1pt}
         Method & Acc & Runtime \\
        \Xcline{1-1}{0.4pt}
        \Xhline{1pt}
        FixMatch\cite{4}  & 40.31  & \textbf{0.08}   \\
        FlexMatch\cite{15}  & 45.50  & 0.14  \\
        \Xcline{1-1}{0.4pt}
        \Xhline{1pt}
        STUC-SSIC  & \textbf{46.89}  & \textbf{0.08}  \\
        \Xhline{1pt}
    \end{tabular}
\end{wraptable}In addition,  semi-supervised image classification methods all have longer training time on datasets like CIFAR-100, which has more number of classifications. We can see that our proposed method improves the accuracy of the model while reducing the training time of the model from Table 2. This is mainly due to the design of the unreliable sample contrastive loss, this contrastive loss utilizes unreliable samples with confidence below the threshold, mines the discriminative information of unreliable samples, involves all the unlabeled data in the training, and improves the utilization of unlabeled data, which accelerates the convergence speed of the model.
\subsection{Ablation Study}
\subsubsection{Impact of different components.}
To investigate the impact of two different components in STUC-SSIC (SATPL,USCL), we test STUC-SSIC with different combinations of two components on CIFAR-10 with four labels per class. To better analyze the performance, follow LaSSL~\cite{27}, we introduce quantity and quality of pseudo-labels. “Quantity” refers to the amount of high-confidence pseudo-labels, calculated by the ratio of the number of high-confidence predictions to the total number of unlabeled samples. “Quality” measures how many high-confidence predictions are consistent to ground-truth labels, which can be obtained by using the real labels from CIFAR-10.
\par It can be seen from Table 3 that each component matters compared to the vanilla version. And that the best classification performance of the model is achieved when the two modules are used jointly. From Table 3, it can be seen that the STUC-SSIC-V2 method using the adaptive threshold pseudo-labeling strategy has a higher number and quality of pseudo-labels and a higher accuracy of the model relative to the STUC-SSIC-V1 method, which is mainly due to the way of dynamically changing thresholds in STUC-SSIC-V2, where the thresholds are low at the beginning of the training period, and more samples will produce pseudo-labels. In the late stage of training, the model's learning ability is stronger, at which time only samples with high confidence will produce pseudo-labels, improving the quality of pseudo-labels.

\subsubsection{Impact of the self-adaptive threshold pseudo-labeling.}
\begin{figure*}[t]
	\centering
	\begin{subfigure}{0.325\linewidth}
		\centering
		\includegraphics[width=0.9\linewidth]{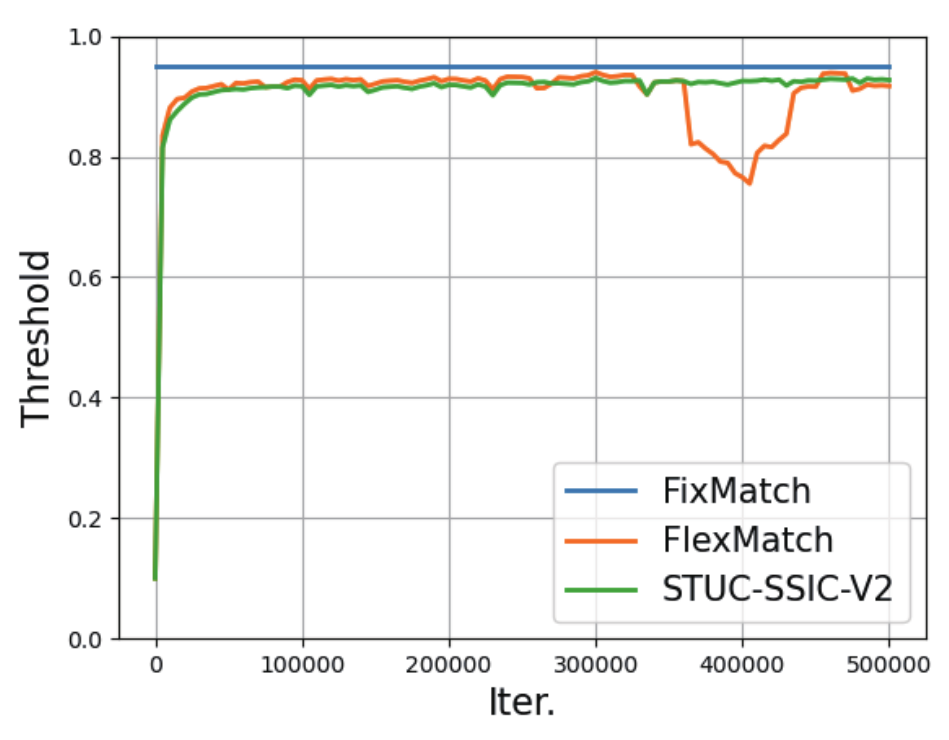}
		\caption{Confidence threshold}
		\label{2a}
	\end{subfigure}
	\centering
	\begin{subfigure}{0.325\linewidth}
		\centering
		\includegraphics[width=0.9\linewidth]{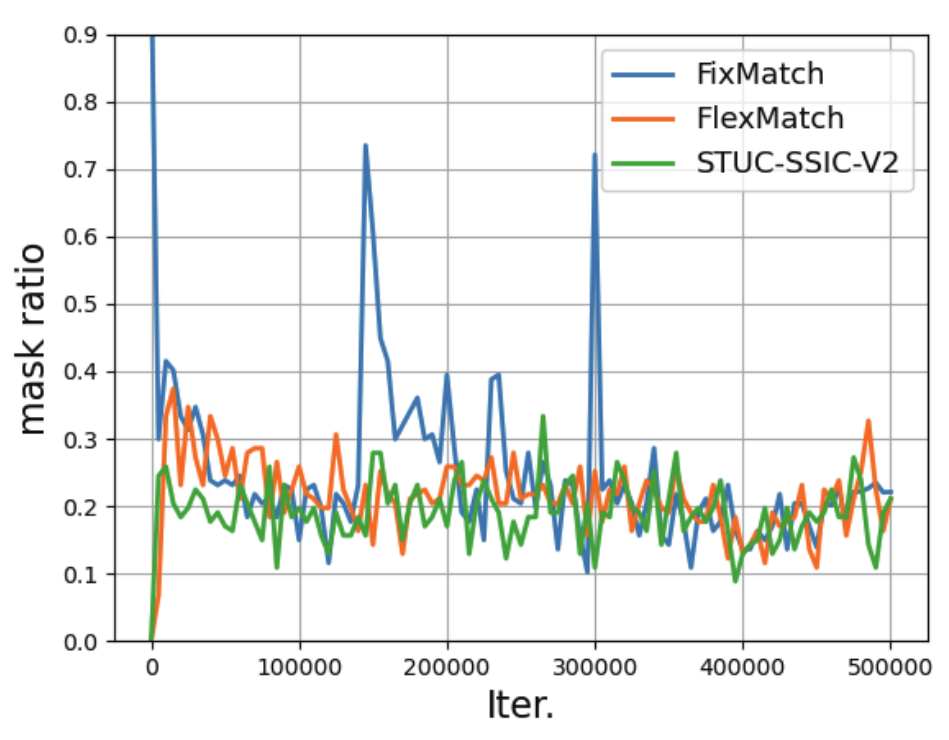}
		\caption{Mask ratio}
		\label{2b}
	\end{subfigure}
        \begin{subfigure}{0.325\linewidth}
		\centering
		\includegraphics[width=0.9\linewidth]{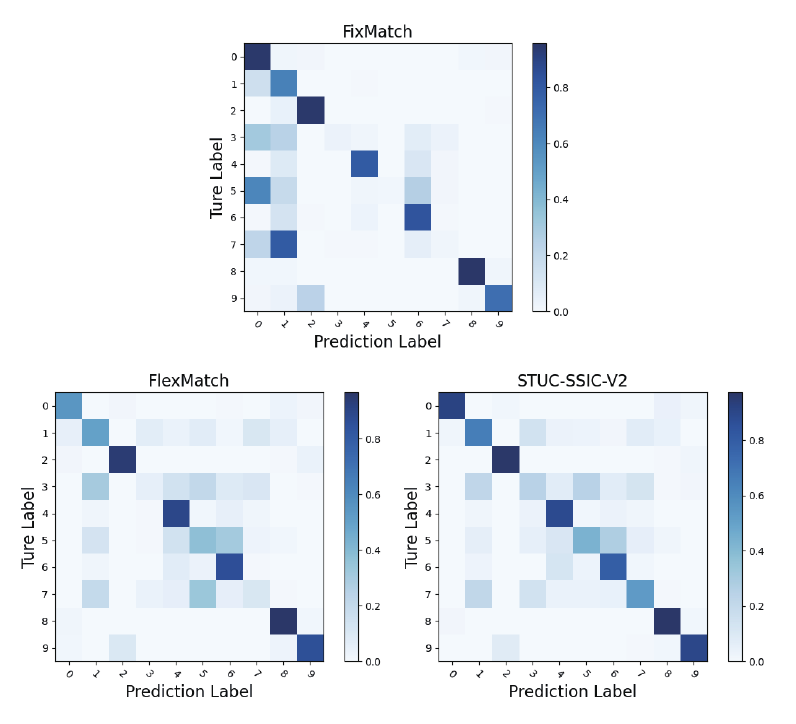}
		\caption{Confusion matrix}
		\label{2c}
	\end{subfigure}
	\caption{Ablation Study of SATPL on STL-10 with 40 labels, compared to previous methods. (a) Class-average confidence threshold; (b) Mask ratio; (c) Confusion matrix, where the fading color of diagonal elements refers to the disparity of the accuracy.}
	\label{fig2}
\end{figure*}

To investigate the validity of SATPL further, we performed an ablation study on the STL-10 with 40  labeles and solely SATPL added to FixMatch. Comparison of the FixMatch method, which uses only a fixed threshold of 0.95, and the FlexMatch method, which uses only dynamic global thresholds. As shown in Fig.\ref{fig2}, where (a) is the class-average confidence threshold, and (b) is the ratio of the number of unlabeled samples screened out to the total number of unlabeled samples, and (c) is the confusion matrix. From Fig. \ref{2a} and Fig. \ref{2b}, it can be seen that the threshold change of the STUC-SSIC-V2 method is consistent with the previous theoretical analysis, and in the early stage of training, STUC-SSIC-V2 has a lower threshold value relative to the FixMatch and FlexMatch methods, and the rate of unlabeled samples 
\begin{table}[t]
    \centering
    \renewcommand{\arraystretch}{1.2}
    \label{tbl:table3}
    \caption{Ablation study(\%) of different components on CIFAR-10 with 40 labels.}
    \begin{tabular}{ccc|ccc}
        \Xhline{1pt}
         Method & SATPL & USCL & Quant & Qual & Acc \\
        \Xcline{1-1}{0.4pt}
        \Xhline{1pt}
        FixMatch\cite{4} & $\times$ & $\times$ & 82.62  & 81.57 & 85.85  \\
        STUC-SSIC-V1 & $\times$ & $\surd$ & 95.72  & 91.25 & 90.81 \\
        STUC-SSIC-V2 & $\surd$ & $\times$ & 96.14  & 92.00 & 91.18  \\
        \Xcline{1-1}{0.4pt}
        \Xhline{1pt}
        \textbf{STUC-SSIC} & $\surd$ & $\surd$ & \textbf{96.77}  & \textbf{92.42} & \textbf{91.71}  \\
        \Xhline{1pt}
    \end{tabular}
\end{table}screened out is lower, so that the utilization rate of unlabeled data increases, thus speeding up the convergence of the model. As the learning ability of the model increases, the STUC-SSIC-V2 threshold is gradually increased to a higher value, alleviating the problem of confirmation bias, improving the quality of pseudo-labels, increasing the classification accuracy of the model, and yielding better classification accuracy (as shown in Fig. \ref{2c}).
\subsubsection{Impact of the different thresholding EMA decay $\lambda$.}\begin{wraptable}{r}{0.45\textwidth}
    \centering
    \label{tbl:table4}
    \caption{Ablation Study of different thresholding EMA decay $\lambda$ on CIFAR-10 with 40 labels.}
    \begin{tabular}{c|c}
        \Xhline{1pt}
         Thresholding EMA decay & Acc(\%) \\
        \Xcline{1-1}{0.4pt}
        \Xhline{1pt}
        0.9  & 91.67   \\
        0.99  & 91.69   \\
        0.999  & \textbf{91.71}   \\
        0.9999  & 91.75   \\
        \Xhline{1pt}
    \end{tabular}
    \vspace{-4mm}
\end{wraptable}We validate the effect of EMA decay parameter $\lambda$ on CIFAR-10 with 40 labels in Table 4, it can be observed that the results of the different $\lambda$ are close, which indicates that STUC-SSIC is robust to $\lambda$ and it is not recommended to use too large a value as it would prevent the updating of the global and local thresholds. The value was set to 0.999 for all experiments.
\subsubsection{Impact of the different $\varepsilon_{1}$ and $\varepsilon_{2}.$}
\begin{wrapfigure}{r}[0cm]{0pt}
    \includegraphics[width=6cm]{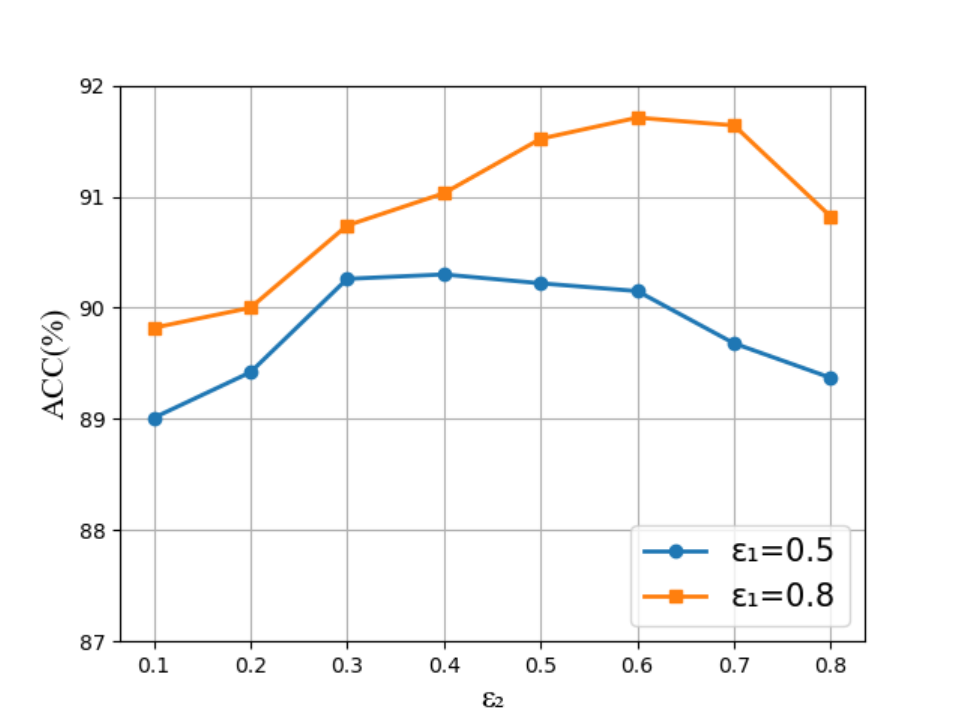}
    \caption{Ablation study of different $\varepsilon_{1}$ and $\varepsilon_{2}$ on CIFAR-10 with 40 labels.}
    \vspace{-0.5cm}
    \label{fig4}
\end{wrapfigure}The contrastive loss proposed in our method introduces two threshold parameters for constructing positive and negative samples, in order to study their effects on the classification performance of the model, we conduct ablation experiments on CIFAR-10 datasets with 40 labels, as shown in Fig.\ref{fig4}. In general, the strong augmentation image is too different from the original image,  so the threshold $\varepsilon_{2}$ for screening the strong augmentation similarity is smaller than the $\varepsilon_{1}$ one. In order to ensure that the screened samples are similar enough, so often set a larger threshold value, in this experiment, the threshold $\varepsilon_{1}$ value is set to 0.5 as well as 0.8, and let the threshold $\varepsilon_{2}$ vary in the range of 0.1 to 0.8. As can be seen from the result, the test performance achieve the best when $\varepsilon_{1}$=0.8, $\varepsilon_{2}$=0.6.
\section{Conclusion}
In this paper, we have proposed a novel semi-supervised learning method STUC-SSIC, which combined with self adaptive threshold pseudo-labeling (SATPL) and unreliable sample contrastive loss (USCL), effectively improves the model’s performance. SATPL adjusts in an adaptive manner to generate more reliable pseudo-labels. In addition, unlike previous approaches, we do not discard samples below the threshold, we propose USCL, which can mine the discriminative information in low-confidence samples by learning the similarities and differences between sample features, enabling the model to utilize all unlabeled samples and speeding up the convergence of the model. Experiment results show that STUC-SSIC can effectively improve pseudo-labels generations in terms of quantity and quality, resulting in better performance over other SSL approaches. Besides, in an open environment, how to utilize contrastive learning for image classification is one of our future work. 
\subsubsection{Acknowledgment.}
This work is supported by the National Natural Science Foundation of China (Nos.62276138, 62076135, 61876087).
%
%
%
%

\end{document}